\documentclass[10pt,twocolumn,letterpaper]{article}
\usepackage[pagenumbers]{cvpr}

\usepackage{xcolor}         %
\usepackage{graphicx}
\usepackage{multirow}
\usepackage{colortbl}
\usepackage{tabularx}
\usepackage[pagebackref=true,breaklinks=true,colorlinks,bookmarks=false]{hyperref}
\usepackage{subcaption}
\usepackage{amsmath} 
\usepackage{amssymb}            
\usepackage{bm}
\usepackage{bbm}

\usepackage{algorithm}
\usepackage{algpseudocode}
\usepackage{marvosym}
\usepackage{inconsolata}

\makeatletter
\newcommand*\bigcdot{\mathpalette\bigcdot@{.5}}
\newcommand*\bigcdot@[2]{\mathbin{\vcenter{\hbox{\scalebox{#2}{$\m@th#1\bullet$}}}}}
\makeatother

\definecolor{cvprblue}{rgb}{0.21,0.49,0.74}
\definecolor{c2}{HTML}{FBD9BD}
\definecolor{c3}{HTML}{fe793d}
\definecolor{c4}{HTML}{eedeb0}
\definecolor{pp}{HTML}{BC7FCD}
\definecolor{bb}{HTML}{CDE8E5}
\definecolor{rouse}{rgb}{0.981,0.961,0.941}

\title{
Beyond Restoration: Content–Degradation Factor Unfolding for Real-World Concealed Object Segmentation
}

\author{Rihan Zhang, Fengyang Xiao, Yuhan Chen, and Sina Farsiu$^{*}$\\
Duke University\\
{\small $^{*}$Corresponding Author, Contact: \href{mailto:rihan.zhang@duke.edu}{rihan.zhang@duke.edu}}}
\hypersetup{pdfauthor={Rihan Zhang, Fengyang Xiao, Yuhan Chen, Sina Farsiu}}

\begin{document}

\twocolumn[{
\maketitle
\vspace{-12mm}

\begin{center}
\includegraphics[width=\textwidth]{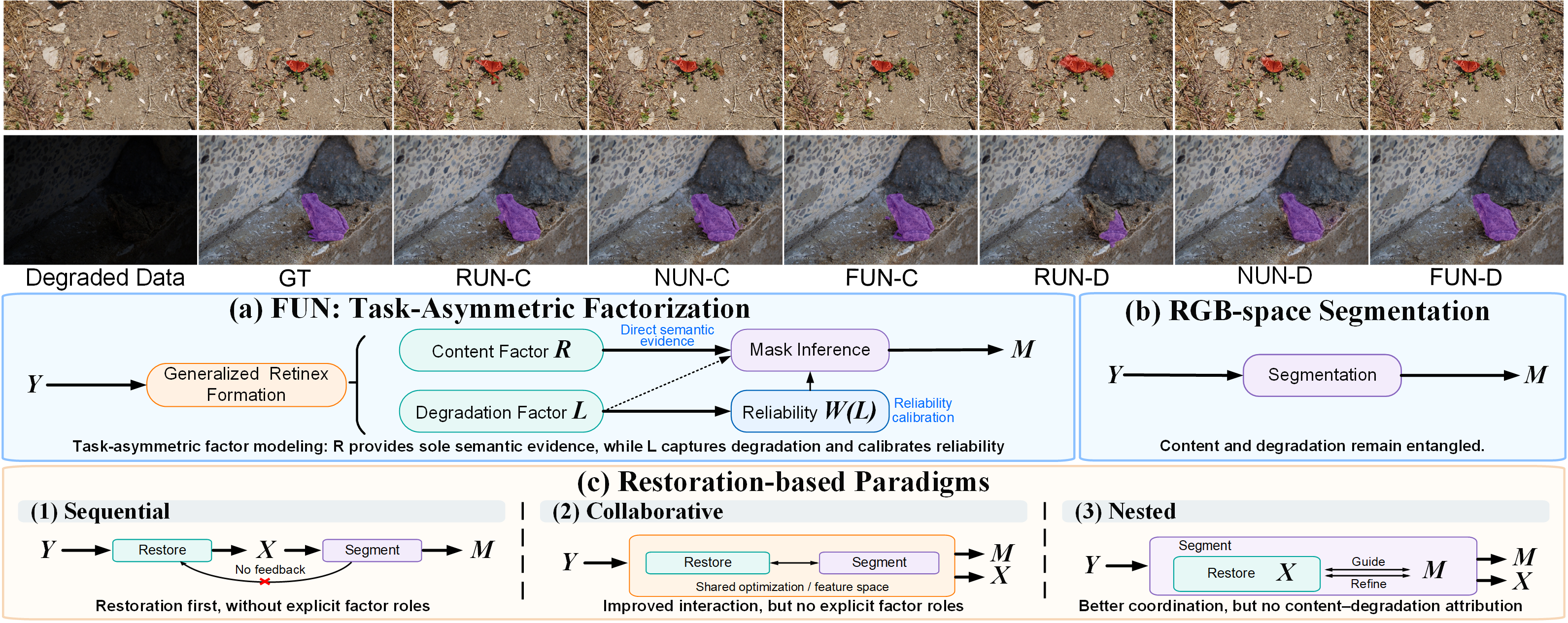}
\vspace{-6mm}
\captionof{figure}{
\textbf{Motivation and qualitative comparisons of FUN.}
\textbf{Top:} Qualitative COS results under clean (``-C'') and degraded (``-D'') conditions.
Red or purple masks are overlaid on the corresponding clean images for visual clarity.
\textbf{Bottom:} Comparison of (a) FUN, (b) standard RGB-space segmentation, and
(c) representative sequential, collaborative, and nested restoration-based segmentation paradigms.
}
\label{Fig:ArchitectureCompare}
\end{center}

}]
\begin{abstract}
Reliable visual perception under degradation requires distinguishing task-relevant content from degradation-induced variations, especially in concealed object segmentation (COS), where foreground cues are inherently weak. 
The central challenge is not merely improving visibility,
but identifying reliable content evidence amid
degradation-induced variations.
RGB-space segmentation often entangles content and degradation cues, while restoration-based methods primarily focus on image recovery and task interaction. This motivates explicit factorization to distinguish content evidence from degradation effects and assess its reliability. Retinex provides a spatially aligned basis for this separation but remains illumination-centric. We therefore extend it to content–degradation formation with task-asymmetric factor roles.
The Factor Unfolding Network (FUN) uses the content factor $R$ as the sole source of direct semantic evidence for segmentation, while the degradation factor $L$ captures degradation-related variations and generates a spatial reliability map to calibrate this evidence. A unified energy couples observation formation with reliability-weighted foreground--background separation, and its alternating updates yield a coarse-to-fine network without clean-image supervision. In this way, FUN factorizes degradation for perception rather than removing it. 
Extensive experiments across clean and degraded conditions and diverse segmentation tasks demonstrate robust performance and generalization. Code will be released. 
\end{abstract}
\newcommand{\FUNcompactmath}{%
  \setlength{\abovedisplayskip}{2pt}%
  \setlength{\belowdisplayskip}{2pt}%
  \setlength{\abovedisplayshortskip}{0pt}%
  \setlength{\belowdisplayshortskip}{2pt}%
  \setlength{\jot}{2pt}%
}
\newsavebox{\FUNmathbox}
\newcommand{\FUNfitmath}[1]{%
  \sbox{\FUNmathbox}{$\displaystyle #1$}%
  \ifdim\wd\FUNmathbox>\dimexpr\columnwidth-3em\relax
    \resizebox{\dimexpr\columnwidth-3em\relax}{!}{\usebox{\FUNmathbox}}%
  \else
    \usebox{\FUNmathbox}%
  \fi
}
\FUNcompactmath
\AtBeginEnvironment{equation}{\FUNcompactmath}
\AtBeginEnvironment{equation*}{\FUNcompactmath}
\AtBeginEnvironment{align}{\FUNcompactmath}
\AtBeginEnvironment{align*}{\FUNcompactmath}

\section{Introduction}
\label{sec:introduction}
Reliable visual perception under degradation requires distinguishing task-relevant content from degradation-induced variations. This challenge is particularly acute in concealed object segmentation (COS), where objects already share similar appearances with their surroundings~\cite{fan2020camouflaged,fan2021concealed,xiao2024survey,He2023Camouflaged,he2023strategic,he2023weaklysupervised,he2025segment}. Under low light, haze, blur, low resolution, or their mixtures, weak object cues become further entangled with degradation responses. 
The fundamental difficulty is therefore not merely
improving visibility, but modeling how concealed-object
cues and degradation-related variations jointly shape
local observations and affect the reliability of content
evidence for segmentation.  

As illustrated in Fig.~\ref{Fig:ArchitectureCompare}, existing methods mainly follow two paradigms. RGB-space segmentors infer content and degradation implicitly from the same mixed representation~\cite{He2023Camouflaged,he2025segment,pang2022zoom}, making a genuine object cue difficult to distinguish from a degradation artifact. Restoration-based methods instead introduce an enhanced image before or during segmentation. Sequential approaches compress degradation uncertainty into a single restored result before semantic reasoning~\cite{xia2023diffir,he2023reti}, whereas collaborative approaches improve task interaction~\cite{he2025run,he2025reversible}, and nested optimization can further alleviate gradient interference~\cite{he2026nun}. 
Nevertheless, these restoration-based methods primarily address how restoration and segmentation should interact. 
Despite their different information flows, explicitly
modeling task-asymmetric roles for content and degradation
remains underexplored across these paradigms.
The qualitative comparisons in Fig.~\ref{Fig:ArchitectureCompare} further illustrate the difficulty of preserving weak object cues under degradation.
Addressing this ambiguity requires an explicit representation in which content-related and degradation-related contributions can assume distinct roles. 
Retinex~\cite{land1971lightness} provides a useful structural starting point by separating an intrinsic factor \(R\) from an environment factor \(L\) that modulates its observable appearance. 
RIDE~\cite{he2026ride} demonstrates the effectiveness
of Retinex-based decomposition for COS on clean data.
However, conventional Retinex remains illumination-centric
and has limited expressiveness for heterogeneous
degradations.
More importantly, factor decomposition alone does not
establish task-asymmetric roles for segmentation,
nor does it specify how degradation-related variations
should calibrate the spatial reliability of content evidence.

To this end, we propose the Factor Unfolding Network (FUN),
built upon a task-asymmetric generalized Retinex formulation
that extends reflectance--illumination decomposition to
content--degradation factorization. A content-dominant factor $R$ and a degradation-dominant factor $L$ jointly explain the degraded observation through nonlinear content-side and degradation-side formation operators. More importantly, the two factors are assigned asymmetric roles during inverse estimation: $R$ is optimized to preserve concealed-object-discriminative content and serves as the sole direct semantic evidence for segmentation, whereas $L$ accounts for complementary degradation-related variations and induces $W(L)$ to calibrate the spatial reliability of that evidence without directly predicting the mask. Thus, the factors remain Retinex-consistent in formation but task-asymmetric in optimization. We therefore factorize degradation for perception rather than remove it before perception.
FUN realizes this principle through a unified optimization model over $R$, $L$, the segmentation mask $M$, and the content-space background $B_c$. An observation-formation term constrains $R$ and $L$ to explain the degraded input, while a reliability-weighted term performs foreground--background separation in the content space. We derive alternating updates for these variables and unfold them into a coarse-to-fine network. Each stage refines the content and degradation explanations, estimates $W(L)$, and uses it to calibrate mask inference in $R$. The framework is trained end-to-end without clean-image supervision.

Our contributions are summarized as follows:
\begin{itemize}
    \item We present, to the best of our knowledge, the first
    task-asymmetric generalized Retinex formulation for degraded COS.
    Learned formation operators jointly explain the observation,
    while $R$ provides semantic evidence and $L$ captures
    degradation-related variations and calibrates
    evidence reliability through $W(L)$.

    \item We establish an optimization model over
$R$, $L$, $M$, and $B_c$, coupling observation formation
with reliability-weighted content-space separation.
Alternating updates unfold into a
coarse-to-fine network for mask inference.

    \item Extensive experiments demonstrate competitive
    performance on clean COS and superior results under
    synthetic degradations, with robust transfer to real-world
    and unseen conditions. Paradigm comparisons and factor-role
    analyses further validate the proposed formulation.
\end{itemize}

\section{Related Work}
\noindent\textbf{Concealed object segmentation.}
Early COS methods, such as SINet~\cite{fan2020camouflaged}, ZoomNet~\cite{pang2022zoom}, and FEDER~\cite{He2023Camouflaged}, extract discriminative cues from similar foregrounds and backgrounds, primarily on clean inputs. To handle degradation, RUN~\cite{he2025run} integrates restoration into unfolded segmentation, while NUN~\cite{he2026nun} coordinates both through nested optimization. RIDE~\cite{he2026ride} uses Retinex-informed decomposition to reveal complementary cues for concealed-object perception. However, preserving weak concealed-object cues under
diverse and unseen degradations remains challenging:
degradation can obscure discriminative content and
reduce its reliability, particularly near object boundaries.

\noindent\textbf{Retinex factorization and deep unfolding networks.}
Retinex~\cite{land1971lightness} decomposes images into intrinsic reflectance and illumination, offering a spatially aligned formulation.
However, conventional Retinex is illumination-centric, limiting its applicability to diverse degradations.
Deep unfolding networks (DUNs) translate iterative optimization into trainable stages and have been applied to restoration~\cite{fang2024real,he2025unfoldir} and COS~\cite{he2025run,he2026nun}. Existing methods predominantly unfold restoration or segmentation objectives, without jointly imposing nonlinear content--degradation formation and reliability-weighted foreground--background separation. FUN assigns asymmetric roles: $R$ alone provides direct semantic evidence, while $L$ retains degradation-related variations and modulates evidence reliability via $W(L)$.

\begin{figure*}[t]
\setlength{\abovecaptionskip}{0cm}
	\centering
	\includegraphics[width=\linewidth]{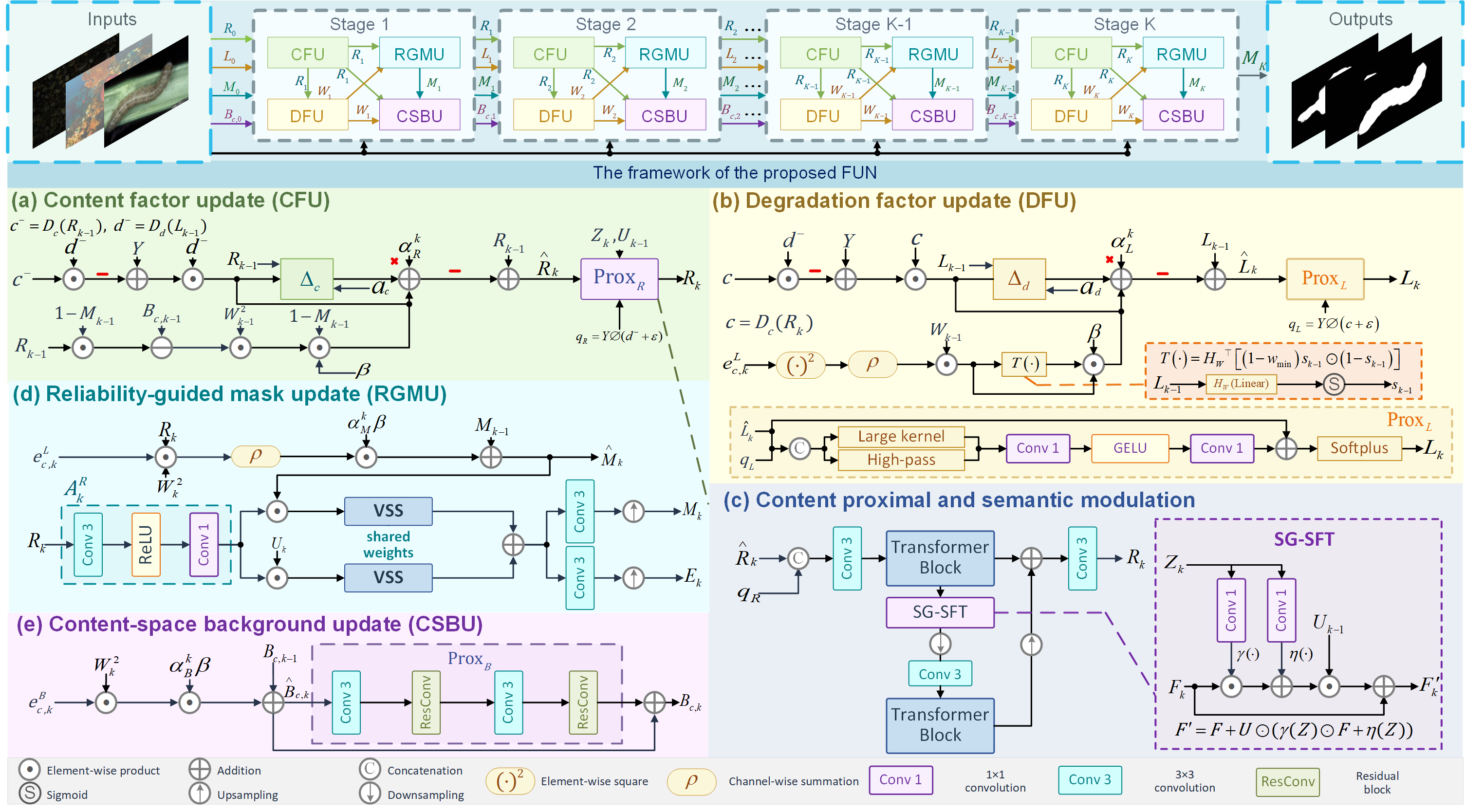}%
	\caption{Framework of our proposed FUN.
    }
	\label{fig:Framework}
	\vspace{-7mm}
\end{figure*}

\section{Task-Asymmetric Generalized Retinex}
\label{sec:formulation}

\subsection{Generalized Retinex Formation}
\label{sec:factor_formation}

Conventional Retinex models image formation through
pointwise reflectance--illumination multiplication,
capturing illumination modulation.
However, heterogeneous degradations, such as
blur-induced spatial mixing, are not fully captured
by this pointwise formulation.
We therefore introduce learnable spatial and affine
transformations for flexible
multiplicative recomposition.
Given a degraded observation $Y$, we introduce
content- and degradation-dominant factors $R$ and $L$,
whose observation-space responses are generated
by nonlinear operators $\mathcal D_c$ and $\mathcal D_d$:
\begin{equation}
Y=\mathcal D_c(R)\odot\mathcal D_d(L),
\label{eq}
\end{equation}
where $\odot$ denotes element-wise multiplication.
Eq.~\eqref{eq} retains a
Retinex-inspired multiplicative formation structure,
where degradation-side responses modulate content-side
responses to explain the observation.
The factors are task-oriented, without assuming
unique physical disentanglement between content
and degradation.

\subsection{Reliability-Weighted Factor-Space Separation}
\label{sec:content_separation}

Observation formation constrains how $R$ and $L$ explain $Y$, but does not
determine their task roles. Rather than requiring $R$ to reproduce a
perceptually clean image, we optimize it to preserve content discriminative
for concealed-object segmentation. We introduce a soft segmentation mask
$M$ and a content-space background $B_c$, and formulate
\begin{equation}
R=R\odot M+B_c,
\label{eq:content_separation}
\end{equation}
where $R\odot M$ represents the target-related content selected by $M$,
while $B_c$ accounts for the remaining background.
Since degradation affects the reliability of content evidence,
we derive a bounded spatial reliability map from $L$:
\begin{equation}
\FUNfitmath{W(L)=w_{\min}+(1-w_{\min})
\sigma\!\left(H_W(L)\right),\quad 0<w_{\min}<1,}
\label{eq:reliability_map}
\end{equation}
where $H_W$ is a learned linear $1\times1$ projection and $\sigma(\cdot)$
is the sigmoid function. Larger values of $W(L)$ indicate more reliable
content evidence, while $w_{\min}$ prevents complete suppression of
the separation constraint. Thus, $R$ serves as the sole direct source of 
semantic evidence for segmentation, whereas $L$ retains degradation-related
variations and uses $W(L)$ to calibrate the spatial reliability of that
evidence without directly predicting $M$.

\subsection{Unified Factor-Space Energy}
\label{sec:factor_energy}

We combine observation formation and reliability-weighted content separation
in the following energy:
\begin{equation}
\FUNfitmath{\begin{aligned}
&\mathcal E(R,L,M,B_c)=\frac{1}{2}
\left\|Y-\mathcal D_c(R)\odot\mathcal D_d(L)\right\|_2^2
\\
&\quad+\frac{\beta}{2}
\left\|W(L)\odot(R-R\odot M-B_c)\right\|_2^2
\\
&\quad+\lambda_R\Omega_R(R)+\lambda_L\Omega_L(L)
+\lambda_M\Omega_M(M)+\lambda_B\Omega_B(B_c),
\end{aligned}}
\label{eq:factor_energy}
\end{equation}
where $\beta$ balances the two data terms, and
$\Omega_R$, $\Omega_L$, $\Omega_M$, and $\Omega_B$
encode factor-specific priors.
The formation term constrains $R$ and $L$ to explain
the degraded observation, with $W(L)$ weighting
content-space separation.
The separation term couples the mask and background
estimates, while stagewise mask supervision and
learned priors provide the semantic constraints.

\section{Factor Unfolding Network}
\label{sec:network}

As illustrated in \cref{fig:Framework}, we unfold the alternating proximal-gradient solver of Eq.~\eqref{eq:factor_energy} into $K=4$ coarse-to-fine stages. Each stage sequentially updates $R_k$, $L_k$, $M_k$, and $B_{c,k}$, with $W_k=W(L_k)$ computed to guide subsequent updates. The formation operators, adjoint-correction modules, and proximal cores are shared across stages, while step sizes and scale-alignment projections are stage-specific. 

We implement the generalized formation operators in
Eq.~\eqref{eq} through lightweight residual and affine mappings:
\begin{equation}
\FUNfitmath{\mathcal D_i(X)=a_i\odot[X+\mathcal C_i(X)]+b_i,
\quad(i,X)\in\{(c,R),(d,L)\},}
\label{eq:residual_formation}
\end{equation}
where $\mathcal C_i$ employs two $3\times3$ convolutions with an
intermediate ReLU and a $3\!\rightarrow\!32\!\rightarrow\!3$
channel configuration to capture local spatial interactions.
The learned, input-independent affine coefficients $a_i$ and $b_i$
provide response scaling and shifting, respectively.
For an observation-space gradient $V$, the Jacobian-transpose
action of each formation operator decomposes into an analytic
direct path and a nonlinear residual path:
\begin{equation}
\begin{aligned}
J_{\mathcal D_i}(X)^\top V
&=a_i\odot V+J_{\mathcal C_i}(X)^\top(a_i\odot V)\\
&\approx a_i\odot V+\Delta_i(a_i\odot V;X).
\end{aligned}
\label{eq:formation_adjoint}
\end{equation}
Here, $\Delta_i$ is a lightweight state-conditioned gated
convolution that approximates the nonlinear residual-path
adjoint. Its architecture is detailed in Sec.~A.1 of the
supplementary material. The content-side and degradation-side modules
$\Delta_c$ and $\Delta_d$ have independent parameters and
are shared across unfolding stages. The direct path and other data-term gradients are
computed analytically, while the priors are handled
by learned proximal mappings.

\noindent\textbf{Content Factor Update (CFU)}.
We initialize the content factor around the observation:
\begin{equation}
R_0
=
Y+\delta_R\tanh\!\left(
H_R^{\mathrm{init}}\!\left(H_{\mathrm{init}}(Y)\right)
\right),
\label{eq:r_initialization}
\end{equation}
where $H_{\mathrm{init}}$ is a shared two-layer $3\times3$
convolutional stem with ReLU activations
($3\!\rightarrow\!32\!\rightarrow\!32$),
and $H_R^{\mathrm{init}}$ is a $3\times3$ projection
to three channels. The scalar $\delta_R>0$ controls
the initialization perturbation.

At stage $k$, let $W_{k-1}=W(L_{k-1})$. Holding $L_{k-1}$,
$M_{k-1}$, and $B_{c,k-1}$ fixed, the content gradient step is
\begingroup
\small
\begin{equation}
\FUNfitmath{\begin{aligned}
&\widehat R_k
=R_{k-1}-\alpha_R^k\Big\{
a_c\odot e_{Y,k}^{R}\odot\mathcal D_d(L_{k-1})\\
&\quad+\Delta_c\!\left(
a_c\odot e_{Y,k}^{R}\odot\mathcal D_d(L_{k-1});
R_{k-1}
\right)\\
&\quad+\beta(\mathbf 1-M_{k-1})\odot W_{k-1}^2
\odot e_{c,k}^{R}
\Big\},
\end{aligned}}
\label{eq:r_gradient}
\end{equation}
\endgroup
where
$e_{Y,k}^{R}=\mathcal D_c(R_{k-1})\odot\mathcal D_d(L_{k-1})-Y$
and
$e_{c,k}^{R}=R_{k-1}-R_{k-1}\odot M_{k-1}-B_{c,k-1}$
are the observation-formation and content-separation residuals, respectively.
The first two terms in Eq.~\eqref{eq:r_gradient} implement the
analytic-plus-correction adjoint in Eq.~\eqref{eq:formation_adjoint}, while the
last term is the closed-form separation gradient.

The content prior is imposed through
\begin{equation}
R_k
=\operatorname{Prox}_R\!\left(
\widehat R_k;\,
Y\oslash[\mathcal D_d(L_{k-1})+\epsilon],
Z_k,U_{k-1}
\right),
\label{eq:r_prox}
\end{equation}
where $U_{k-1}=(1-W_{k-1})/(1-w_{\min})$
denotes normalized unreliability.
A lightweight two-scale Restormer~\cite{zamir2022restormer}
encodes the concatenation of $\widehat R_k$ and the
formation-derived quotient. Following~\cite{he2026nun},
$Z_k$ is selected in a deep-to-shallow manner from
backbone features extracted once from $Y$.
To adaptively inject segmentation-oriented features
according to content reliability, we introduce
a segmentation-guided spatial feature transform (SG-SFT):
{\small
\begin{equation}
\FUNfitmath{\operatorname{SG\text{-}SFT}(F_k,Z_k,U_{k-1})
=F_k+U_{k-1}\odot
\left[\gamma(Z_k)\odot F_k+\eta(Z_k)\right],}
\label{eq:sgsft}
\end{equation}}
where $F_k$ is an intermediate Restormer feature,
and $\gamma(\cdot)$ and $\eta(\cdot)$ are lightweight
convolutional projections generating spatial affine
parameters from $Z_k$. The unreliability map
$U_{k-1}$ adaptively controls feature correction.
The quotient provides observation-derived conditioning
without requiring a clean-image target.

\noindent\textbf{Degradation Factor Update (DFU)}.
The degradation factor is initialized from the shared shallow stem as
\begin{equation}
L_0
=\exp\!\left(
\delta_L\tanh\!\left(
H_L^{\mathrm{init}}\!\left(H_{\mathrm{init}}(Y)\right)
\right)
\right),
\label{eq:l_initialization}
\end{equation}
where $H_L^{\mathrm{init}}$ is a $3\times3$ projection
to three channels, and $\delta_L>0$ controls the
initialization range. The exponential centers $L_0$
around one and ensures positivity.

\begin{table*}[t]
\centering
\renewcommand{\arraystretch}{0.88}
\setlength{\aboverulesep}{0.7pt}
\setlength{\belowrulesep}{0.7pt}
\captionsetup{position=top,skip=2pt}
\caption{Quantitative comparison on clean COD benchmarks.
FUN is trained separately from its degraded-data counterpart.}
\label{tab:clean_cod}
\setlength{\tabcolsep}{1.35mm}
\resizebox{\textwidth}{!}{%
\begin{tabular}{l|c|cccc|cccc|cccc|cccc}
\toprule
\multirow{2}{*}{Methods}
& \multirow{2}{*}{Backbone}
& \multicolumn{4}{c|}{CHAMELEON}
& \multicolumn{4}{c|}{CAMO}
& \multicolumn{4}{c|}{COD10K}
& \multicolumn{4}{c}{NC4K} \\
\cline{3-18}
& &
\cellcolor{gray!25}$M\downarrow$ &
\cellcolor{gray!25}$F_\beta\uparrow$ &
\cellcolor{gray!25}$E_\phi\uparrow$ &
\cellcolor{gray!25}$S_\alpha\uparrow$ &
\cellcolor{gray!25}$M\downarrow$ &
\cellcolor{gray!25}$F_\beta\uparrow$ &
\cellcolor{gray!25}$E_\phi\uparrow$ &
\cellcolor{gray!25}$S_\alpha\uparrow$ &
\cellcolor{gray!25}$M\downarrow$ &
\cellcolor{gray!25}$F_\beta\uparrow$ &
\cellcolor{gray!25}$E_\phi\uparrow$ &
\cellcolor{gray!25}$S_\alpha\uparrow$ &
\cellcolor{gray!25}$M\downarrow$ &
\cellcolor{gray!25}$F_\beta\uparrow$ &
\cellcolor{gray!25}$E_\phi\uparrow$ &
\cellcolor{gray!25}$S_\alpha\uparrow$ \\
\midrule

CamoFocus~\cite{khan2024camofocus}
& PVTv2-B4
& 0.023 & 0.869 & 0.953 & 0.906
& 0.044 & 0.861 & 0.924 & 0.870
& 0.022 & 0.818 & 0.931 & 0.868
& 0.031 & 0.862 & 0.932 & 0.886 \\

CamoDiff~\cite{sun2025conditional}
& PVTv2-B4
& 0.022 & 0.868 & 0.952 & 0.908
& 0.042 & 0.853 & 0.936 & 0.878
& 0.019 & 0.815 & 0.943 & 0.883
& 0.028 & 0.858 & 0.942 & 0.895 \\

RUN~\cite{he2025run}
& PVTv2-B4
& 0.021 & 0.877 & 0.958 & 0.916
& 0.045 & 0.861 & 0.934 & 0.877
& 0.021 & 0.810 & 0.941 & 0.878
& 0.030 & 0.868 & 0.940 & 0.892 \\

NUN~\cite{he2026nun}
& PVTv2-B4
& 0.021 & 0.888 & 0.963 & 0.918
& 0.042 & 0.868 & 0.939 & 0.885
& 0.018 & \textbf{0.839} & 0.948 & 0.890
& \textbf{0.027} & 0.880 & 0.947 & 0.897 \\

RIDE~\cite{he2026ride}
& PVTv2-B4
& 0.020 & 0.886 & 0.965 & 0.920
& 0.041 & 0.868 & 0.940 & 0.887
& 0.018 & 0.838 & 0.950 & 0.889
& \textbf{0.027} & 0.883 & \textbf{0.948} & 0.898 \\

\rowcolor{c2!20}
\textbf{FUN (Ours)}
& PVTv2-B4
& \textbf{0.019} & \textbf{0.893} & \textbf{0.968} & \textbf{0.924}
& \textbf{0.040} & \textbf{0.869} & \textbf{0.941} & \textbf{0.888}
& \textbf{0.017} & 0.834 & \textbf{0.956} & \textbf{0.907}
& \textbf{0.027} & \textbf{0.884} & \textbf{0.948} & \textbf{0.906} \\

\midrule

VSCode (adapted)~\cite{luo2024vscode}
& DINOv2-L
& 0.020 & 0.882 & 0.954 & 0.915
& 0.043 & 0.860 & 0.931 & 0.882
& 0.020 & 0.818 & 0.935 & 0.880
& 0.030 & 0.867 & 0.940 & 0.895 \\

C3Net~\cite{jan2025c3net}
& DINOv2-L
& 0.016 & 0.905 & 0.968 & 0.927
& 0.031 & 0.892 & 0.953 & 0.904
& 0.016 & 0.852 & 0.961 & 0.898
& 0.024 & 0.894 & 0.958 & 0.913 \\

NUN~\cite{he2026nun}
& DINOv2-L
& 0.015 & 0.918 & 0.975 & 0.931
& 0.028 & 0.908 & 0.965 & 0.911
& 0.014 & 0.880 & 0.972 & 0.905
& 0.020 & 0.913 & 0.967 & 0.920 \\

RIDE~\cite{he2026ride}
& DINOv2-L
& \textbf{0.014} & \textbf{0.920} & 0.978 & 0.933
& 0.028 & 0.911 & 0.967 & 0.912
& \textbf{0.013} & \textbf{0.887} & \textbf{0.977} & 0.906
& 0.019 & 0.914 & \textbf{0.969} & 0.921 \\

\rowcolor{c2!20}
\textbf{FUN (Ours)}
& DINOv2-L
& 0.015 & 0.906 & \textbf{0.979} & \textbf{0.937}
& \textbf{0.027} & \textbf{0.915} & \textbf{0.968} & \textbf{0.922}
& \textbf{0.013} & 0.871 & 0.973 & \textbf{0.919}
& \textbf{0.018} & \textbf{0.916} & \textbf{0.969} & \textbf{0.930} \\

\bottomrule
\end{tabular}}
\vspace{1pt}
\vspace{-6MM}
\end{table*}

Using the latest $R_k$ while holding $M_{k-1}$ and $B_{c,k-1}$ fixed, the
degradation gradient step is
\begingroup
\small
\begin{equation}
\FUNfitmath{\begin{aligned}
&\widehat L_k
=L_{k-1}-\alpha_L^k\Big\{
a_d\odot e_{Y,k}^{L}\odot\mathcal D_c(R_k)\\
&\quad+\Delta_d\!\left(
a_d\odot e_{Y,k}^{L}\odot\mathcal D_c(R_k);
L_{k-1}
\right)\\
&\quad+\beta J_W(L_{k-1})^\top
\left[W_{k-1}\odot\rho(\left(e_{c,k}^{L}\right)^2)\right]
\Big\},
\end{aligned}}
\label{eq:l_gradient}
\end{equation}
\endgroup
where
$e_{Y,k}^{L}=\mathcal D_c(R_k)\odot
\mathcal D_d(L_{k-1})-Y$ and
$e_{c,k}^{L}=R_k-R_k\odot M_{k-1}-B_{c,k-1}$
are the formation and separation residuals,
respectively, and $\rho(\cdot)$ denotes channel-wise
summation. The first two terms implement the
adjoint in Eq.~\eqref{eq:formation_adjoint},
while the last propagates the reliability-weighted
separation gradient through $W(L)$.

Because $H_W$ in Eq.~\eqref{eq:reliability_map} is linear, the reliability
adjoint is evaluated exactly:
{\small
\begin{equation}
\FUNfitmath{\begin{aligned}
&J_W(L_{k-1})^\top
\left[W_{k-1}\odot\rho(\left(e_{c,k}^{L}\right)^2)\right]\\
&\quad=H_W^\top\!\Big[
(1-w_{\min})s_{k-1}\odot(1-s_{k-1})\\
&\qquad\odot W_{k-1}\odot\rho(\left(e_{c,k}^{L}\right)^2)
\Big],
\end{aligned}}
\label{eq:w_adjoint}
\end{equation}}
where $s_{k-1}=\sigma(H_W(L_{k-1}))$. The operator $H_W^\top$ reuses the
forward projection weights and introduces no additional reverse-mapping
parameters.

The degradation prior is subsequently imposed by
\begin{equation}
L_k
=\operatorname{Prox}_L\!\left(
\widehat L_k;\,
Y\oslash[\mathcal D_c(R_k)+\epsilon]
\right).
\label{eq:l_prox}
\end{equation}
The quotient provides observation-derived conditioning
under the latest content estimate. The lightweight degradation proximal employs a large-kernel spatial branch to capture broader degradation patterns and a high-pass residual branch to provide complementary local corrections. Their fused features predict a residual refinement, followed by Softplus to ensure $L_k>0$. Architectural details are provided in Sec.~A.1 of the supplementary material. It receives no backbone feature $Z_k$.
Finally, the reliability map is refreshed as
$W_k=W(L_k)$ for subsequent updates.

\noindent\textbf{Reliability-Guided Mask Update (RGMU)}.
The mask is initialized as $M_0=\mathbf 0$.
Using the updated content factor $R_k$ and reliability map $W_k$,
the mask gradient step is
\begin{equation}
\widehat M_k
=M_{k-1}+\alpha_M^k\beta\,
\rho\!\left(R_k\odot W_k^2\odot e_{c,k}^{L}\right).
\label{eq:m_gradient}
\end{equation}
The mask gradient is computed analytically without additional reverse mappings.

Let $U_k=(1-W_k)/(1-w_{\min})$. The mask is refined through
\begin{equation}
(M_k,E_k)
=\operatorname{Prox}_M\!\left(
\widehat M_k;\,
A_R^k(R_k),U_k
\right),
\label{eq:m_prox}
\end{equation}
where $A_R^k$ aligns $R_k$ to the stage resolution.
The mask proximal employs two weight-shared 
VSS paths~\cite{Liu2024vmamba}: one aggregates content
evidence guided by $\widehat M_k$, while the other
focuses on unreliable regions indicated by $U_k$.
Their fused features predict the mask $M_k$ and
auxiliary edge $E_k$, which are upsampled to the input
resolution. Semantic evidence is drawn exclusively
from $R_k$, while $L_k$ influences mask inference
through reliability modulation rather than direct
feature injection. The backbone feature $Z_k$ is
excluded from $\operatorname{Prox}_M$, and $E_k$
is used only for auxiliary supervision.

\noindent\textbf{Content-Space Background Update (CSBU)}.
The content-space background is initialized as $B_{c,0}=\mathbf 0$. After
updating $M_k$, define
$e_{c,k}^{B}=R_k-R_k\odot M_k-B_{c,k-1}$. Its closed-form gradient step is
\begin{equation}
\widehat B_{c,k}
=B_{c,k-1}+\alpha_B^k\beta W_k^2\odot e_{c,k}^{B}.
\label{eq:b_gradient}
\end{equation}
The background prior is then imposed by
\begin{equation}
B_{c,k}=\operatorname{Prox}_B(\widehat B_{c,k}).
\label{eq:b_prox}
\end{equation}
The lightweight background proximal contains two residual convolutional
blocks~\cite{he2023HQG} and predicts a local refinement. Because
$R_k$, $M_k$, and $W_k$ already enter the explicit update, they are not
reintroduced into $\operatorname{Prox}_B$; neither $Z_k$ nor an additional
segmentation prediction is used.

\section{Training Objective}
\label{sec:training_objective}

We supervise all $K=4$ stages using weighted BCE and IoU losses
for segmentation, together with an auxiliary Dice boundary loss~\cite{he2025run}.
To enforce the proposed factorization, we introduce two
Charbonnier-based consistency losses: observation formation
constrains the reconstruction of $Y$, while reliability-weighted
content separation regularizes the relationship among $R$, $M$,
and $B_c$. The reliability map is detached in the separation
loss to prevent trivial weight suppression, while its gradients
remain active in the unfolded updates.
The overall objective is
\begin{equation}
\FUNfitmath{\mathcal L_{\mathrm{total}}
=\sum_{k=1}^{K}\omega_k\big(
\mathcal L_{\mathrm{seg}}^k
+\lambda_{\mathrm{bnd}}\mathcal L_{\mathrm{bnd}}^k
+\lambda_{\mathrm{obs}}\mathcal L_{\mathrm{obs}}^k
+\lambda_{\mathrm{sep}}\mathcal L_{\mathrm{sep}}^k
\big),}
\label{eq:overall_loss}
\end{equation}
where $\omega_k=2^{k-K}$. FUN is trained without clean-image
supervision or ground truth for $L$ and $B_c$.
Detailed loss formulations are provided in Sec.~A.2.

\begin{table*}[t]
\centering
\renewcommand{\arraystretch}{0.88}
\setlength{\aboverulesep}{0.7pt}
\setlength{\belowrulesep}{0.7pt}
\captionsetup{position=top,skip=2pt}
\caption{Quantitative comparison on COD benchmarks under different
synthetic degradation settings.}
\label{tab:degraded_cod}
\setlength{\tabcolsep}{1mm}
\resizebox{\textwidth}{!}{%
\begin{tabular}{l|c|cccc|cccc|cccc|cccc}
\toprule
\multirow{2}{*}{Methods}
& \multirow{2}{*}{Backbone}
& \multicolumn{4}{c|}{CHAMELEON}
& \multicolumn{4}{c|}{CAMO}
& \multicolumn{4}{c|}{COD10K}
& \multicolumn{4}{c}{NC4K} \\
\cline{3-18}
&
& \cellcolor{gray!25}$M\downarrow$
& \cellcolor{gray!25}$F_\beta\uparrow$
& \cellcolor{gray!25}$E_\phi\uparrow$
& \cellcolor{gray!25}$S_\alpha\uparrow$
& \cellcolor{gray!25}$M\downarrow$
& \cellcolor{gray!25}$F_\beta\uparrow$
& \cellcolor{gray!25}$E_\phi\uparrow$
& \cellcolor{gray!25}$S_\alpha\uparrow$
& \cellcolor{gray!25}$M\downarrow$
& \cellcolor{gray!25}$F_\beta\uparrow$
& \cellcolor{gray!25}$E_\phi\uparrow$
& \cellcolor{gray!25}$S_\alpha\uparrow$
& \cellcolor{gray!25}$M\downarrow$
& \cellcolor{gray!25}$F_\beta\uparrow$
& \cellcolor{gray!25}$E_\phi\uparrow$
& \cellcolor{gray!25}$S_\alpha\uparrow$ \\
\midrule

\multicolumn{18}{c}{\textbf{Single Degradation: Blur}} \\
\midrule

RUN~\cite{he2025run}
& ResNet-50
& 0.091 & 0.620 & 0.801 & 0.768
& 0.141 & 0.581 & 0.680 & 0.675
& 0.071 & 0.530 & 0.766 & 0.732
& 0.095 & 0.637 & 0.793 & 0.768 \\

RIDE~\cite{he2026ride}
& ResNet-50
& 0.088 & 0.655 & 0.797 & 0.776
& 0.137 & 0.628 & 0.735 & 0.718
& 0.069 & 0.537 & 0.778 & 0.737
& 0.093 & 0.663 & 0.792 & 0.770 \\

NUN~\cite{he2026nun}
& ResNet-50
& 0.068 & 0.720 & 0.810 & 0.796
& 0.129 & 0.632 & 0.697 & 0.690
& 0.062 & 0.561 & 0.778 & 0.745
& 0.082 & 0.692 & 0.796 & 0.777 \\

\rowcolor{c2!20}
\textbf{FUN (Ours)}
& ResNet-50
& \textbf{0.055} & \textbf{0.721} & \textbf{0.879} & \textbf{0.821}
& \textbf{0.113} & \textbf{0.661} & \textbf{0.785} & \textbf{0.726}
& \textbf{0.053} & \textbf{0.593} & \textbf{0.832} & \textbf{0.760}
& \textbf{0.072} & \textbf{0.699} & \textbf{0.847} & \textbf{0.788} \\

\midrule

\multicolumn{18}{c}{\textbf{Single Degradation: Low-light}} \\
\midrule

RUN~\cite{he2025run}
& ResNet-50
& 0.057 & 0.729 & 0.841 & 0.813
& 0.124 & 0.579 & 0.737 & 0.657
& 0.063 & 0.539 & 0.785 & 0.684
& 0.089 & 0.643 & 0.795 & 0.713 \\

RIDE~\cite{he2026ride}
& ResNet-50
& 0.066 & 0.691 & 0.842 & 0.794
& 0.113 & 0.658 & 0.768 & 0.720
& 0.060 & 0.561 & 0.804 & 0.737
& 0.075 & 0.697 & 0.829 & 0.777 \\

NUN~\cite{he2026nun}
& ResNet-50
& 0.054 & 0.736 & 0.851 & 0.824
& 0.112 & 0.634 & 0.754 & 0.724
& 0.055 & 0.608 & 0.795 & 0.751
& 0.072 & 0.714 & 0.826 & 0.794 \\

\rowcolor{c2!20}
\textbf{FUN (Ours)}
& ResNet-50
& \textbf{0.042} & \textbf{0.839} & \textbf{0.912} & \textbf{0.872}
& \textbf{0.093} & \textbf{0.693} & \textbf{0.782} & \textbf{0.748}
& \textbf{0.045} & \textbf{0.688} & \textbf{0.847} & \textbf{0.783}
& \textbf{0.067} & \textbf{0.741} & \textbf{0.842} & \textbf{0.805} \\

\midrule

\multicolumn{18}{c}{\textbf{Single Degradation: Haze}} \\
\midrule

RUN~\cite{he2025run}
& ResNet-50
& 0.067 & 0.652 & 0.769 & 0.748
& 0.128 & 0.547 & 0.681 & 0.636
& 0.059 & 0.528 & 0.726 & 0.731
& 0.079 & 0.678 & 0.754 & 0.764 \\

RIDE~\cite{he2026ride}
& ResNet-50
& 0.063 & 0.687 & 0.802 & 0.789
& 0.123 & 0.565 & 0.695 & 0.672
& 0.056 & 0.570 & 0.765 & 0.738
& 0.075 & 0.692 & 0.798 & 0.776 \\

NUN~\cite{he2026nun}
& ResNet-50
& 0.059 & 0.706 & 0.815 & 0.805
& 0.120 & 0.574 & 0.704 & 0.685
& 0.053 & 0.598 & 0.780 & 0.743
& 0.072 & 0.699 & 0.811 & 0.782 \\

\rowcolor{c2!20}
\textbf{FUN (Ours)}
& ResNet-50
& \textbf{0.044} & \textbf{0.783} & \textbf{0.906} & \textbf{0.869}
& \textbf{0.094} & \textbf{0.716} & \textbf{0.827} & \textbf{0.777}
& \textbf{0.047} & \textbf{0.620} & \textbf{0.852} & \textbf{0.790}
& \textbf{0.059} & \textbf{0.745} & \textbf{0.874} & \textbf{0.828} \\

\midrule
\multicolumn{18}{c}{
\textbf{Combined Degradation: Blur, Low-light, and Haze}
} \\
\midrule

RUN~\cite{he2025run}
& ResNet-50
& 0.104 & 0.562 & 0.688 & 0.676
& 0.153 & 0.488 & 0.575 & 0.570
& 0.072 & 0.474 & 0.666 & 0.640
& 0.106 & 0.555 & 0.675 & 0.660 \\

RIDE~\cite{he2026ride}
& ResNet-50
& 0.100 & 0.591 & 0.697 & 0.686
& 0.149 & 0.545 & 0.615 & 0.605
& 0.074 & 0.473 & 0.692 & 0.661
& 0.109 & 0.590 & 0.686 & 0.669 \\

NUN~\cite{he2026nun}
& ResNet-50
& 0.081 & 0.634 & 0.709 & 0.689
& 0.139 & 0.519 & 0.592 & 0.587
& 0.068 & 0.500 & 0.680 & 0.648
& 0.094 & 0.627 & 0.717 & 0.688 \\

\rowcolor{c2!20}
\textbf{FUN (Ours)}
& ResNet-50
& \textbf{0.078} & \textbf{0.676} & \textbf{0.790} & \textbf{0.741}
& \textbf{0.127} & \textbf{0.584} & \textbf{0.715} & \textbf{0.651}
& \textbf{0.060} & \textbf{0.556} & \textbf{0.785} & \textbf{0.703}
& \textbf{0.088} & \textbf{0.654} & \textbf{0.788} & \textbf{0.720} \\

\cmidrule{1-18}

RIDE~\cite{he2026ride}
& PVTv2-B4
& 0.086 & 0.688 & 0.792 & 0.776
& 0.129 & 0.664 & 0.751 & 0.738
& 0.065 & 0.612 & 0.804 & 0.763
& 0.090 & 0.711 & 0.804 & 0.786 \\

NUN~\cite{he2026nun}
& PVTv2-B4
& 0.066 & 0.704 & 0.839 & 0.793
& 0.109 & 0.690 & 0.782 & 0.745
& 0.056 & 0.605 & 0.824 & 0.767
& 0.071 & 0.723 & 0.843 & 0.800 \\

\rowcolor{c2!20}
\textbf{FUN (Ours)}
& PVTv2-B4
& \textbf{0.057} & \textbf{0.725} & \textbf{0.849} & \textbf{0.806}
& \textbf{0.102} & \textbf{0.707} & \textbf{0.796} & \textbf{0.748}
& \textbf{0.044} & \textbf{0.661} & \textbf{0.853} & \textbf{0.788}
& \textbf{0.065} & \textbf{0.745} & \textbf{0.850} & \textbf{0.802} \\

\midrule
\multicolumn{18}{c}{\textbf{Unseen Degradation: Low Resolution}} \\
\midrule

RIDE~\cite{he2026ride}
& PVTv2-B4
& 0.129 & 0.587 & 0.735 & 0.725
& 0.167 & 0.589 & 0.726 & 0.718
& 0.090 & 0.533 & 0.763 & 0.728
& 0.119 & 0.631 & 0.772 & 0.756 \\

NUN~\cite{he2026nun}
& PVTv2-B4
& 0.116 & 0.546 & 0.704 & 0.671
& 0.161 & 0.564 & 0.662 & 0.654
& 0.084 & 0.488 & 0.739 & 0.684
& 0.119 & 0.587 & 0.728 & 0.698 \\

\rowcolor{c2!20}
\textbf{FUN (Ours)}
& PVTv2-B4
& \textbf{0.068} & \textbf{0.684} & \textbf{0.826} & \textbf{0.785}
& \textbf{0.116} & \textbf{0.659} & \textbf{0.791} & \textbf{0.728}
& \textbf{0.054} & \textbf{0.604} & \textbf{0.829} & \textbf{0.765}
& \textbf{0.073} & \textbf{0.701} & \textbf{0.840} & \textbf{0.787} \\
\bottomrule
\end{tabular}}
\vspace{-4MM}
\end{table*}

\section{Experiments}
\label{sec:experiment}

\noindent\textbf{Experimental setup}. FUN is implemented in PyTorch and trained on a single NVIDIA H200 GPU using Adam with momentum terms $(0.9, 0.999)$. Following the common practice~\cite{he2026nun,he2025run}, we incorporate multi-level deep features extracted from encoder-shaped backbones into our framework. All input images are resized to $352\times352$. The batch size is set to 16, and the initial learning rate is $1\times10^{-4}$, decreased by a factor of 0.1 every 80 epochs. The number of unfolding stages $K$ is fixed at 4. Architectural and hyperparameter details are provided in Sec.~A.1.

\subsection{COS tasks with clean data}
Datasets and metrics are detailed in Sec.~B.1.
For clean data, FUN fixes $W=1$ and replaces SG-SFT by
unit-gated standard SFT, preserving backbone feature injection,
content--degradation factorization, and unfolding structure.

\noindent\textbf{Camouflaged object detection}.
Following NUN~\cite{he2026nun}, we evaluate FUN using two backbones: PVT V2~\cite{wang2022pvt} and DINOv2~\cite{oquab2023dinov2}. As shown in \cref{tab:clean_cod}, FUN achieves competitive performance across four benchmarks with both backbones.

\begin{table*}[t]
\begingroup
\setlength{\abovecaptionskip}{0cm}
	\centering
	\includegraphics[width=\linewidth]{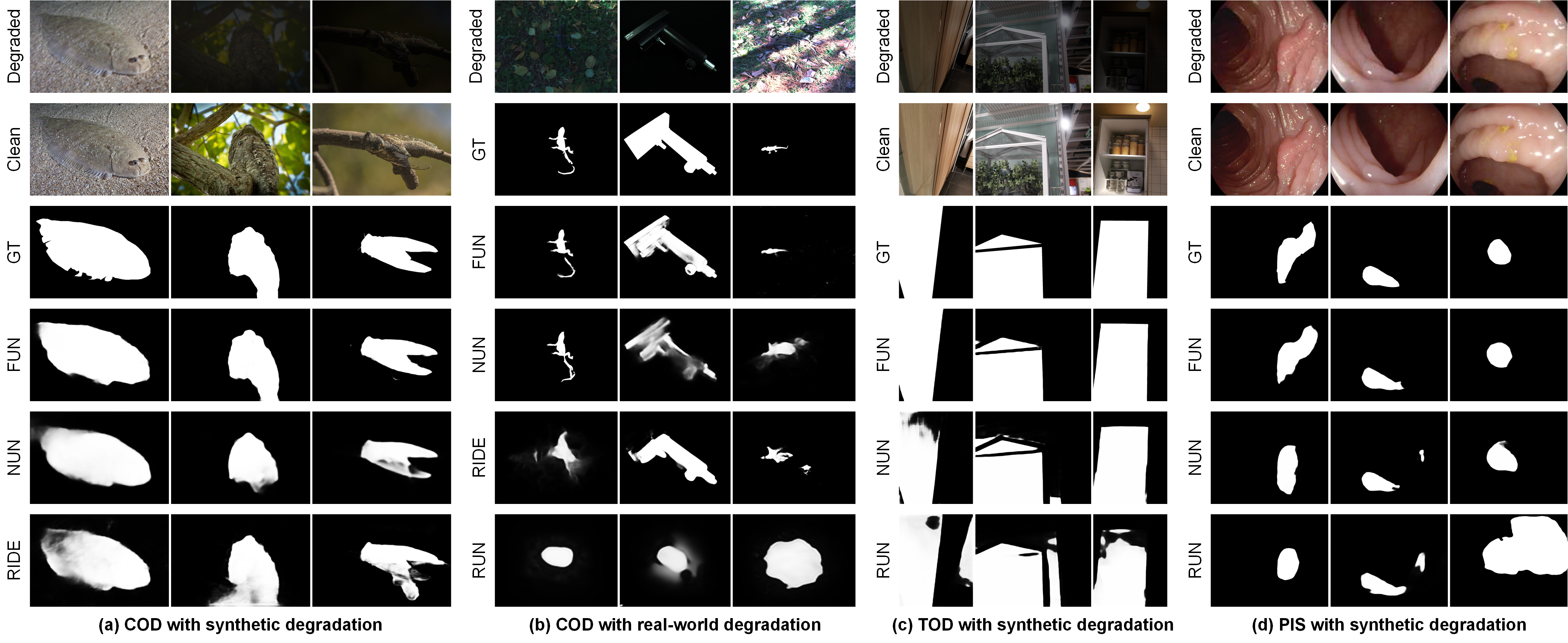}%
	\captionof{figure}{Visualization on COS tasks with real-world and synthetic degradation.
    }
	\label{fig:Visualization}
	
\endgroup
\par\vspace{2pt}

\centering
\renewcommand{\arraystretch}{0.88}
\setlength{\aboverulesep}{0.7pt}
\setlength{\belowrulesep}{0.7pt}
\captionsetup{position=top,skip=2pt}
\caption{Learning and architecture ablations on COD10K under combined degradation.}
\label{tab:components}

\resizebox{\textwidth}{!}{
\setlength{\tabcolsep}{0.35mm}
\begin{tabular}{c|cc|cc|cc|cccc|c}
\toprule

\multirow{2}{*}{Metrics}
& \multicolumn{2}{c|}{Supervision}
& \multicolumn{2}{c|}{Reliability}
& \multicolumn{2}{c|}{Reverse Mapping}
& \multicolumn{4}{c|}{Proximal Realization}
& \cellcolor{c2!20} FUN
\\
\cline{2-11}

& w/o $\mathcal L_{\rm obs}$
& w/o $\mathcal L_{\rm sep}$
& $U_{\rm spa}\!\rightarrow\!U_{\rm img}$
& $U\nrightarrow\mathcal P_M$
& $\mathcal A_c\!\rightarrow\!\Delta_c$
& $\mathcal A_d\!\rightarrow\!\Delta_d$
& $\mathcal S_U\!\rightarrow\!\mathcal S$
& $\mathcal M_V\!\rightarrow\!\mathcal M_C$
& $\mathcal L_{MF}\!\rightarrow\!\mathcal L_C$
& $\mathcal B_R\!\rightarrow\!\mathcal B_T$
& \cellcolor{c2!20} Ours
\\
\midrule

$M\downarrow$
& 0.070
& 0.069
& 0.064
& 0.063
& 0.068
& 0.067
& 0.064
& 0.065
& 0.064
& 0.061
& \cellcolor{c2!20}\textbf{0.060}
\\

$F_\beta\uparrow$
& 0.468
& 0.477
& 0.531
& 0.524
& 0.496
& 0.504
& 0.524
& 0.518
& 0.527
& 0.548
& \cellcolor{c2!20}\textbf{0.556}
\\

$E_\phi\uparrow$
& 0.713
& 0.707
& 0.760
& 0.768
& 0.727
& 0.736
& 0.764
& 0.751
& 0.758
& 0.780
& \cellcolor{c2!20}\textbf{0.785}
\\

$S_\alpha\uparrow$
& 0.660
& 0.667
& 0.688
& 0.694
& 0.678
& 0.671
& 0.687
& 0.688
& 0.690
& 0.699
& \cellcolor{c2!20}\textbf{0.703}
\\

\bottomrule
\end{tabular}}
\par\vspace{3pt}
\begingroup
\fontsize{8}{9}\selectfont
\renewcommand{\arraystretch}{0.90}
\setlength{\tabcolsep}{1.4pt}
\setlength{\aboverulesep}{0.5pt}
\setlength{\belowrulesep}{0.5pt}
\captionsetup{font=small,position=top,skip=1.5pt,justification=centering,singlelinecheck=false}
\begingroup
\setlength{\heavyrulewidth}{0.5pt}
\setlength{\lightrulewidth}{0.3pt}
\begin{minipage}[t]{0.475\textwidth}
\vspace{0pt}
\centering
\caption{Segmentation paradigms on COD10K.}
\label{tab:paradigms}
\begin{tabularx}{\linewidth}{@{}l|c|>{\centering\arraybackslash}X>{\centering\arraybackslash}X>{\centering\arraybackslash}X>{\centering\arraybackslash}X@{}}
\toprule
\rule[-3pt]{0pt}{12pt}Methods & Params(M)
& $M\downarrow$
& $F_\beta\uparrow$
& $E_\phi\uparrow$
& $S_\alpha\uparrow$ \\
\midrule
RGB-Matched
& 29.13 & 0.076 & 0.438 & 0.651 & 0.624 \\
Sequential
& 29.23 & 0.073 & 0.456 & 0.669 & 0.633 \\
RUN~\cite{he2025run}
& 30.41 & 0.072 & 0.474 & 0.666 & 0.640 \\
NUN~\cite{he2026nun}
& 33.17 & 0.068 & 0.500 & 0.680 & 0.648 \\
\rowcolor{c2!20}
FUN
& \textbf{29.01}
& \textbf{0.060}
& \textbf{0.556}
& \textbf{0.785}
& \textbf{0.703} \\
\bottomrule
\end{tabularx}
\end{minipage}
\hfill
\begin{minipage}[t]{0.51\textwidth}
\vspace{0pt}
\centering
\caption{Unfolding ablations on COD10K.}
\label{tab:formulation}
\begin{tabularx}{\linewidth}{@{}l|>{\centering\arraybackslash}X>{\centering\arraybackslash}X>{\centering\arraybackslash}X>{\centering\arraybackslash}X|>{\centering\arraybackslash}X>{\centering\arraybackslash}X>{\centering\arraybackslash}X>{\centering\arraybackslash}X@{}}
\toprule
\rule[-3pt]{0pt}{12pt}Variant & Dual & Form. & Sep. & Rel.
& $M\downarrow$
& $F_\beta\uparrow$
& $E_\phi\uparrow$
& $S_\alpha\uparrow$ \\
\midrule
A
& $\checkmark$ & -- & -- & --
& 0.078 & 0.417 & 0.641 & 0.620 \\[2.025pt]
B
& $\checkmark$ & $\checkmark$ & -- & --
& 0.072 & 0.463 & 0.697 & 0.654 \\[2.025pt]
C
& $\checkmark$ & $\checkmark$ & $\checkmark$ & --
& 0.065 & 0.512 & 0.748 & 0.681 \\[2.025pt]
\rowcolor{c2!20}
FUN
& $\checkmark$ & $\checkmark$ & $\checkmark$ & $\checkmark$
& \textbf{0.060}
& \textbf{0.556}
& \textbf{0.785}
& \textbf{0.703} \\[2.025pt]
\bottomrule
\end{tabularx}
\end{minipage}
\endgroup
\par\vspace{2pt}
\begin{minipage}[t]{\textwidth}
\vspace{0pt}
\centering
\begin{minipage}[t][12pt][t]{\linewidth}
\caption{Functional roles of content and degradation factors
under different degradations.}
\label{tab:factor_roles}
\end{minipage}
\par

\begin{tabularx}{\linewidth}{@{}l|>{\centering\arraybackslash}X>{\centering\arraybackslash}X|>{\centering\arraybackslash}X>{\centering\arraybackslash}X|>{\centering\arraybackslash}X>{\centering\arraybackslash}X>{\centering\arraybackslash}X@{}}
\toprule
\multirow{2}{*}{Degradation}
& \multicolumn{2}{c|}{Free Branches}
& \multicolumn{2}{c|}{Retinex}
& \multicolumn{3}{c}{Full FUN} \\
\cline{2-8}

& $Q_{RY}\uparrow$
& $Q_{RL}\uparrow$
& $Q_{RY}\uparrow$
& $Q_{RL}\uparrow$
& \cellcolor{c2!20}$Q_{RY}\uparrow$
& \cellcolor{c2!20}$Q_{RL}\uparrow$
& \cellcolor{c2!20}$\rho_s(U,d)\uparrow$ \\
\midrule

Blur
& 0.436 & 0.448
& 0.481 & 0.572
& \cellcolor{c2!20}\textbf{0.710}
& \cellcolor{c2!20}\textbf{0.814}
& \cellcolor{c2!20}0.634 \\

Low-light
& 0.492 & 0.519
& 0.728 & 0.812
& \cellcolor{c2!20}\textbf{0.792}
& \cellcolor{c2!20}\textbf{0.883}
& \cellcolor{c2!20}0.744 \\

Haze
& 0.463 & 0.477
& 0.547 & 0.648
& \cellcolor{c2!20}\textbf{0.771}
& \cellcolor{c2!20}\textbf{0.870}
& \cellcolor{c2!20}0.695 \\

\bottomrule
\end{tabularx}
\end{minipage}

\endgroup
\vspace{-6MM}
\end{table*}

\begin{table*}[!t]
\centering
\begingroup

    \centering
    \includegraphics[width=\textwidth]{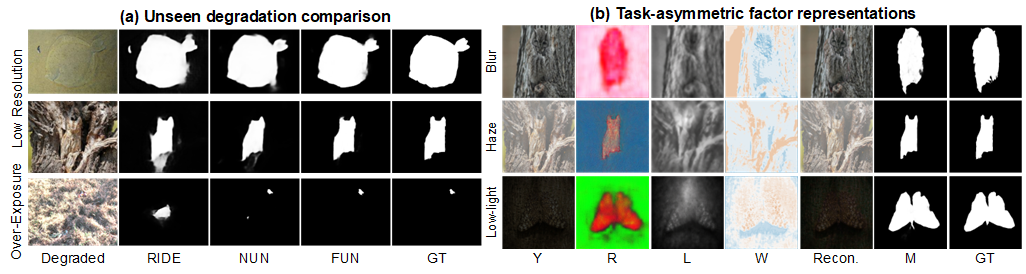}
    \vspace{-8mm}
    \captionof{figure}{\textbf{Unseen Degradations and Factor Visualization.}
(a) Segmentation results of RIDE, NUN, and FUN under unseen degradations.
(b) Observation $Y$, content factor $R$, degradation factor $L$,
reliability $W(L)$, reconstruction $\widetilde{Y}$, mask $M$, and GT. $R$ emphasizes discriminative content, while $L$ captures degradation cues and modulates reliability.
$R$ is shown in RGB; $L$ is jointly min--max normalized across
all channels and pixels and shown in grayscale.
$W(L)$ uses a fixed $[0.1,1]$ color scale
(blue: low reliability; orange: high reliability).
}
    \label{fig:qualitative_factor}
    \vspace{-3mm}

\endgroup
\par\vspace{2pt}

\centering
\newcommand{\FUNheadcenter}[1]{\raisebox{\dimexpr6.55pt-(\height-\depth)/2\relax}[0pt][0pt]{#1}}
\fontsize{8}{9}\selectfont
\renewcommand{\arraystretch}{0.90}
\setlength{\tabcolsep}{1.4pt}
\setlength{\aboverulesep}{0.5pt}
\setlength{\belowrulesep}{0.5pt}
\setlength{\heavyrulewidth}{0.5pt}
\setlength{\lightrulewidth}{0.3pt}
\captionsetup{font=small,position=top,skip=1.5pt,justification=centering,singlelinecheck=false}
\begingroup
\setlength{\heavyrulewidth}{0.8pt}
\setlength{\lightrulewidth}{0.5pt}
\begin{minipage}[t]{0.33\textwidth}
\vspace{0pt}
\centering
\begin{minipage}[t][12pt][t]{\linewidth}
\caption{Ablation of generalized Retinex.}
\label{tab:formation_ablation}
\end{minipage}
\par

\begin{tabularx}{\linewidth}{@{}l|>{\centering\arraybackslash}X>{\centering\arraybackslash}X>{\centering\arraybackslash}X@{}}
\toprule
\multirow{2}{*}{Metrics}
& \multicolumn{3}{c}{Formation Operators} \\
\cmidrule[\arrayrulewidth]{2-4}
& Identity
& Residual
& \cellcolor{c2!20}Full \\
\midrule

$M\downarrow$
& 0.074 & 0.065
& \cellcolor{c2!20}\textbf{0.060} \\[1.0375pt]

$F_\beta\uparrow$
& 0.481 & 0.537
& \cellcolor{c2!20}\textbf{0.556} \\[1.0375pt]

$E_\phi\uparrow$
& 0.714 & 0.772
& \cellcolor{c2!20}\textbf{0.785} \\[1.0375pt]

$S_\alpha\uparrow$
& 0.654 & 0.692
& \cellcolor{c2!20}\textbf{0.703} \\[1.0375pt]

\bottomrule
\end{tabularx}
\end{minipage}\hfill%
\begin{minipage}[t]{0.65\textwidth}
\vspace{0pt}
\centering
\begin{minipage}[t][12pt][t]{\linewidth}
\caption{Factor input and reliability source ablations on COD10K.}
\label{tab:factor_inputs_reliability}
\end{minipage}
\par
\begin{tabularx}{\linewidth}{@{}l|cc|>{\centering\arraybackslash}X>{\centering\arraybackslash}X>{\centering\arraybackslash}X>{\centering\arraybackslash}X@{}}
\toprule
\rule[-3pt]{0pt}{12pt}Variant & Direct Features & Reliability
& $M\downarrow$ & $F_\beta\uparrow$ & $E_\phi\uparrow$ & $S_\alpha\uparrow$ \\
\midrule
L-only & $L$ & $W(L)$
& 0.076 & 0.463 & 0.710 & 0.652 \\

R+L Concat & $[R,L]$ & $W(L)$
& 0.065 & 0.528 & 0.763 & 0.687 \\

$W(Y)$ & $R$ & $W(Y)$
& 0.068 & 0.502 & 0.725 & 0.671 \\

$W(R)$ & $R$ & $W(R)$
& 0.066 & 0.511 & 0.752 & 0.689 \\
\rowcolor{c2!20}
\textbf{Full FUN} & $R$ & $W(L)$ & \textbf{0.060} & \textbf{0.556} & \textbf{0.785} & \textbf{0.703} \\
\bottomrule
\end{tabularx}
\end{minipage}
\endgroup
\par\vspace{2pt}
\begin{minipage}[t]{0.43\textwidth}
\vspace{0pt}
\centering
\setlength{\tabcolsep}{0.6pt}
\caption{Results on real-world degraded COD.}
\label{tab:real_cod}
\begin{tabularx}{\linewidth}{@{}l|>{\centering\arraybackslash}X>{\centering\arraybackslash}X>{\centering\arraybackslash}X|>{\centering\arraybackslash}X>{\centering\arraybackslash}X>{\centering\arraybackslash}X@{}}
\toprule
\multirow{2}{*}{Methods}
& \multicolumn{3}{c|}{PCOD-LQ}
& \multicolumn{3}{c}{MCOD-LQ} \\
\cline{2-7}
\rule[-4pt]{0pt}{11pt}& $E_\phi^{\mathrm{mean}}\mkern-4mu\uparrow$
& $S_\alpha\uparrow$
& $M\downarrow$
& $E_\phi^{\mathrm{mean}}\mkern-4mu\uparrow$
& $S_\alpha\uparrow$
& $M\downarrow$ \\
\midrule
RUN~\cite{he2025run}
& 0.601 & 0.635 & 0.078
& 0.468 & 0.485 & 0.043 \\[2.025pt]
RIDE~\cite{he2026ride}
& 0.627 & 0.660 & 0.064
& 0.492 & 0.509 & 0.038 \\[2.025pt]
NUN~\cite{he2026nun}
& 0.667 & 0.663 & 0.058
& 0.605 & 0.580 & 0.030 \\[2.025pt]
\rowcolor{c2!20}
\textbf{FUN}
& \textbf{0.693} & \textbf{0.708} & \textbf{0.050}
& \textbf{0.628} & \textbf{0.598} & \textbf{0.024} \\[2.025pt]
\bottomrule
\end{tabularx}
\end{minipage}
\hfill
\begin{minipage}[t]{0.275\textwidth}
\vspace{0pt}
\centering
\caption{Results on degraded TOD.}
\label{tab:degraded_tod}
\begin{tabularx}{\linewidth}{@{}l|>{\centering\arraybackslash}X>{\centering\arraybackslash}X>{\centering\arraybackslash}X@{}}
\toprule
\multicolumn{1}{@{}c@{}|}{\rule[-3pt]{0pt}{19.1pt}\FUNheadcenter{Methods}}
& \FUNheadcenter{mIoU$\uparrow$}
& \FUNheadcenter{$F_\beta^{\max}\uparrow$}
& \FUNheadcenter{$M\downarrow$} \\
\midrule
EBLNet~\cite{he2021enhanced}
& 0.786 & 0.873 & 0.108 \\
RFENet~\cite{fan2023rfenet}
& 0.835 & 0.896 & 0.081 \\
RUN~\cite{he2025run}
& 0.862 & 0.920 & 0.070 \\
NUN~\cite{he2026nun}
& 0.877 & 0.928 & 0.062 \\
\rowcolor{c2!20}
\textbf{FUN (Ours)}
& \textbf{0.879}
& \textbf{0.929}
& \textbf{0.059} \\
\bottomrule
\end{tabularx}
\end{minipage}
\hfill
\begin{minipage}[t]{0.275\textwidth}
\vspace{0pt}
\centering
\caption{Results on degraded PIS.}
\label{tab:degraded_pis}
\begin{tabularx}{\linewidth}{@{}l|>{\centering\arraybackslash}X>{\centering\arraybackslash}X>{\centering\arraybackslash}X@{}}
\toprule
\multicolumn{1}{@{}c@{}|}{\rule[-3pt]{0pt}{19.1pt}\FUNheadcenter{Methods}}
& \FUNheadcenter{mDice$\uparrow$}
& \FUNheadcenter{mIoU$\uparrow$}
& \FUNheadcenter{$S_\alpha\uparrow$} \\
\midrule
Polyp-PVT~\cite{dong2023polyp}
& 0.580 & 0.511 & 0.731 \\
MEGANet~\cite{bui2024meganet}
& 0.667 & 0.596 & 0.810 \\
RUN~\cite{he2025run}
& 0.715 & 0.638 & 0.830 \\
NUN~\cite{he2026nun}
& 0.722 & 0.647 & 0.836 \\
\rowcolor{c2!20}
\textbf{FUN (Ours)}
& \textbf{0.725}
& \textbf{0.650}
& \textbf{0.837} \\
\bottomrule
\end{tabularx}
\end{minipage}
\vspace{-6MM}

\end{table*}

\noindent\textbf{Other COS tasks}.
We evaluate FUN on transparent object detection (TOD), polyp image segmentation (PIS), and concealed defect detection (CDD). TOD and PIS use GDD and ETIS, respectively, with PVTv2-B4 following existing settings~\cite{he2025run,he2026nun}. For CDD, following RUN~\cite{he2025run}, the COD-trained PVTv2-B4 model is applied to CDS2K without further adaptation. Results are reported in Tables~A1--A3 in the supplementary material.

\subsection{COS tasks with degraded data}

\noindent\textbf{COD with synthetic degradation}.
We simulate challenging real-world conditions using three representative degradations, \textit{i.e.}, blur, low-light, and haze, together with their combined degradation~\cite{he2023reti,he2025unfoldir}, and retrain all methods under the same corruption settings. Detailed dataset descriptions and degradation protocols are provided in Sec.~B.2. As shown in \cref{tab:degraded_cod,fig:Visualization}, FUN consistently outperforms the competing methods across the four COD benchmarks under blur, low-light, haze, and combined degradation, demonstrating the advantage of explicitly separating content- and degradation-related responses for robust segmentation. The gains remain pronounced under the more challenging combined degradation, further validating the effectiveness of factor-space reasoning when multiple degradations are entangled. Moreover, replacing ResNet-50 with the stronger PVT V2 backbone improves the combined-degradation results, indicating that FUN generalizes well across different encoder architectures. 

\noindent\textbf{COD with real degradation}.
Following NUN~\cite{he2026nun}, we further evaluate FUN on the real-world low-quality COD subsets PCOD-LQ~\cite{wang2023polarization} and MCOD-LQ~\cite{li2025mcod}. The model trained under the combined synthetic degradation is directly evaluated without further fine-tuning or adaptation. As shown in \cref{tab:real_cod,fig:Visualization}, FUN maintains strong segmentation performance on naturally degraded observations, demonstrating effective transfer from synthetic to real-world degradations and further supporting the practical relevance of our synthetic degradation setting. 

\noindent\textbf{Other COS tasks with synthetic degradation}.
Following NUN~\cite{he2026nun}, we further evaluate FUN on degraded transparent object detection and polyp image segmentation. Scenario-specific degradations are applied according to the characteristics of each task (see  Sec.~B.2). As shown in \cref{tab:degraded_tod,tab:degraded_pis,fig:Visualization}, FUN maintains strong performance across these degraded COS scenarios, demonstrating the general applicability of the proposed factor-space formulation beyond degraded COD.

\subsection{Ablation Study}
\label{sec:ablation}
All ablations are conducted on COD10K under combined degradation using the ResNet-50 backbone.

\noindent\textbf{Comparison across segmentation paradigms}.
To assess the task-asymmetric factor-space formulation, we compare FUN with RGB-space, sequential, collaborative, and nested segmentation paradigms. \textit{RGB-Matched} directly predicts masks using the same ResNet-50 encoder and four-level VSS decoder as FUN, with additional residual blocks for parameter matching. \textit{Sequential} employs a two-scale Restormer~\cite{zamir2022restormer} trained with paired clean targets and frozen, followed by the same encoder and decoder. We also include RUN~\cite{he2025run} and NUN~\cite{he2026nun} as representative collaborative and nested restoration--segmentation approaches. Unlike these paradigms, FUN jointly optimizes content, degradation, mask, and background without clean-image supervision. Among the methods in \cref{tab:paradigms},
FUN achieves the best performance with the fewest
parameters, showing that these gains do not require
an increased model parameter budget.

\noindent\textbf{Effect of the unfolding formulation}.
We introduce the formulation under controlled settings. Variant A uses dual branches with generic $L\!\rightarrow\!R$ conditioning but no explicit formation, separation, or reliability modeling. Variant B adds formation-based updates, quotient conditioning, and observation-consistency supervision. Variant C incorporates the background $B_c$ with content-space separation updates and consistency supervision. Variants A--C fix $W=1$ and use standard SFT with a unit gate. FUN learns $W(L)$ to guide explicit updates, mask proximal refinement, and SG-SFT. Thus, A-to-B, B-to-C, and C-to-FUN evaluate observation formation, content separation, and adaptive reliability, respectively. In \cref{tab:formulation}, each addition improves segmentation performance under combined degradation.

\noindent\textbf{Learning and architectural choices}.
\cref{tab:components} evaluates design choices relative to FUN. For supervision, we separately remove $\mathcal L_{\rm obs}$ and $\mathcal L_{\rm sep}$. For reliability, we replace spatial $U$ with its image-wise average ($U_{\rm spa}\rightarrow U_{\rm img}$) or remove its mask-proximal modulation while retaining other pathways. For reverse mappings, $\mathcal A_c\rightarrow\Delta_c$ and $\mathcal A_d\rightarrow\Delta_d$ remove the analytic direct paths from the adjoints. For proximal designs, we replace SG-SFT with SFT ($\mathcal S_U\rightarrow\mathcal S$), VSS with a parameter-matched CNN ($\mathcal M_V\rightarrow\mathcal M_C$), mixed-frequency modeling with CNN ($\mathcal L_{MF}\rightarrow\mathcal L_C$), and the residual background proximal with a Transformer~\cite{he2023reti} ($\mathcal B_R\rightarrow\mathcal B_T$). As shown in \cref{tab:components}, all modifications reduce performance, demonstrating the effectiveness of the designs.

\noindent\textbf{Ablation of factor-role assignment}.
We compare $R$, $L$, and $[R,L]$ as direct inputs to
$\operatorname{Prox}_M$ in Eq.~\eqref{eq:m_prox}, with $W(L)$ fixed.
We further compare reliability sources $W(Y)$,
$W(R)$, and $W(L)$ with matched projections and
corresponding gradient updates.
\cref{tab:factor_inputs_reliability} shows that
the proposed assignment achieves the best performance,
supporting $R$ for direct mask refinement and
$L$ for reliability modulation.

\subsection{Further Analysis}
\label{sec:analysis}
Unfolding depth, efficiency, and limitations/future work
appear in supplementary Secs.~B.3, B.4, and C, respectively.

\noindent\textbf{Performance on unseen degradations}.
FUN, RIDE, and NUN use their respective
combined-degradation checkpoints for unseen evaluation.
FUN demonstrates robust performance on low-resolution
inputs (\cref{tab:degraded_cod}), with qualitative
results under real-world over-exposure
(\cref{fig:qualitative_factor}).

\noindent\textbf{Effect of generalized formation operators}.
We evaluate three formation variants under combined degradation: conventional Retinex ($\mathcal D_i(X)=X$), spatial residual formation ($\mathcal D_i(X)=X+\mathcal C_i(X)$), and full generalized formation with learnable affine coefficients. All other FUN components remain unchanged. As shown in \cref{tab:formation_ablation}, progressive gains show the benefits of nonlinear spatial transformations and affine response modulation to robust segmentation.

\noindent\textbf{Functional roles of the factors}.
We analyze the functional roles of $R$ and $L$ under
blur, low-light, and haze on COD10K images.
Here, \textit{Free Branches} follows Variant A
in \cref{tab:formulation}, while \textit{Retinex} adopts
identity formation (\cref{tab:formation_ablation}).
We measure foreground--background selectivity using
$Q_{RY}$ and $Q_{RL}$, and assess degradation-related
reliability through spatial Spearman correlations
between $U_K$ and degradation-specific severity cues.
Detailed metric definitions and computation protocols
are provided in Sec.~B.5 of the supplementary material.
As shown in \cref{tab:factor_roles}, $R_K$ exhibits
stronger foreground--background selectivity than
both $Y$ and $L_K$, while $U_K$ shows positive
correlations with degradation-related variations. 
Together with \cref{tab:factor_inputs_reliability,fig:qualitative_factor},
these results support using $R$ for discriminative content evidence
and $L$ for spatial reliability modulation.

\section{Conclusion}
We introduced FUN for degraded COS, extending Retinex to
task-asymmetric content--degradation factorization.
$R$ provides the sole direct semantic evidence, while $L$
captures degradation variations and calibrates reliability
through $W(L)$.
Unfolded optimization couples observation formation with
reliability-weighted content separation without clean-image supervision.
Experiments demonstrate robustness across clean, mixed,
and unseen COS conditions. Ablation studies further validate the effectiveness of generalized formation and task-asymmetric factor roles.

\par\clearpage

{
    \small
    \bibliographystyle{ieeenat_fullname}
    \bibliography{main}

@String(CVPR= {IEEE Conf. Comput. Vis. Pattern Recog.})

@String(ICCV= {Int. Conf. Comput. Vis.})

@String(ICLR = {Int. Conf. Learn. Represent.})

@String(IJCAI = {IJCAI})

@String(CVPR  = {CVPR})

@String(ICCV  = {ICCV})

@String(ICLR  = {ICLR})

@article{fan2021concealed,
  title={Concealed object detection},
  author={Fan, Deng-Ping and Ji, Ge-Peng and Cheng, Ming-Ming and Shao, Ling},
  journal={IEEE Trans. Pattern Anal. Mach. Intell.},
  year={2021},
  publisher={IEEE}
}

@article{he2023weaklysupervised,
  title={Weakly-supervised concealed object segmentation with {SAM}-based pseudo labeling and multi-scale feature grouping},
  author={He, Chunming and Li, Kai and Zhang, Yachao and Xu, Guoxia and Tang, Longxiang},
  journal={NeurIPS},
  year={2024}
}

@inproceedings{fan2020camouflaged,
  title={Camouflaged object detection},
  author={Fan, Deng-Ping and Ji, Ge-Peng and Sun, Guolei and Cheng, Ming-Ming and Shen, Jianbing},
  booktitle={CVPR},
  pages={2777--2787},
  year={2020}
}

@inproceedings{He2023Camouflaged,
title={Camouflaged object detection with feature decomposition and edge reconstruction},
author={He, Chunming and Li, Kai and Zhang, Yachao and Tang, Longxiang and Zhang, Yulun},
booktitle={CVPR},
pages={22046--22055},
year={2023}
}

@article{he2023strategic,
 title={Strategic Preys Make Acute Predators: Enhancing Camouflaged Object Detectors by Generating Camouflaged Objects},
 author={He, Chunming and Li, Kai and Zhang, Yachao and Zhang, Yulun and Guo, Zhenhua and Li, Xiu},
 journal={ICLR},
 year={2024}
}

@article{dong2023polyp,
  title={{Polyp-PVT}: Polyp Segmentation with Pyramid Vision Transformers},
  author={Dong, Bo and Wang, Wenhai and Fan, Deng-Ping and Li, Jinpeng and Fu, Huazhu and Shao, Ling},
  journal={CAAI Artif. Intell. Res.},
  volume={2},
  year={2023}
}

@inproceedings{he2021enhanced,
  title={Enhanced boundary learning for glass-like object segmentation},
  author={He, Hao and Li, Xiangtai and Cheng, Guangliang and Shi, Jianping and Tong, Yunhai and Meng, Gaofeng and Prinet, V{\'e}ronique and Weng, LuBin},
  booktitle={ICCV},
  pages={15859--15868},
  year={2021}
}

@inproceedings{fan2023rfenet,
  title={{RFENet}: towards reciprocal feature evolution for glass segmentation},
  author={Fan, Ke and Wang, Changan and Wang, Yabiao and Wang, Chengjie and Yi, Ran and Ma, Lizhuang},
  booktitle={IJCAI},
  pages={717--725},
  year={2023}
}

@inproceedings{luo2024vscode,
  title={{VSCode}: General Visual Salient and Camouflaged Object Detection with {2D} Prompt Learning},
  author={Luo, Ziyang and Liu, Nian and Zhao, Wangbo and Yang, Xuguang and Zhang, Dingwen and Fan, Deng-Ping and Khan, Fahad and Han, Junwei},
  booktitle={CVPR},
  pages={17169--17180},
  year={2024}
}

@inproceedings{khan2024camofocus,
  title={{CamoFocus}: Enhancing Camouflage Object Detection With Split-Feature Focal Modulation and Context Refinement},
  author={Khan, Abbas and Khan, Mustaqeem and Gueaieb, Wail and El Saddik, Abdulmotaleb and De Masi, Giulia and Karray, Fakhri},
  booktitle={WACV},
  pages={1434--1443},
  year={2024}
}

@article{he2023HQG,
  title={{HQG-Net}: Unpaired medical image enhancement with high-quality guidance},
  author={Xu, Guoxia and Yan, Jiangpeng and Tang, Longxiang and Zhang, Yulun},
  journal={IEEE Trans. Neural Networks Learn. Syst.},
  year={2023},
  publisher={IEEE}
}

@article{xiao2024survey,
  title={A Survey of Camouflaged Object Detection and Beyond},
  author={Xiao, Fengyang and Hu, Sujie and Shen, Yuqi and He, Chunming},
  journal={arXiv preprint arXiv:2408.14562},
  year={2024}
}

@inproceedings{fang2024real,
  title={Real-world Image Dehazing with Coherence-based Pseudo Labeling and Cooperative Unfolding Network},
  author={Fang, Chengyu and He, Chunming and Xiao, Fengyang and Zhang, Yulun and Tang, Longxiang and Zhang, Yuelin and Li, Kai and Li, Xiu},
  booktitle={The Thirty-eighth Annual Conference on Neural Information Processing Systems},
  year={2024}
}

@article{wang2022pvt,
  title={{PVT v2}: Improved baselines with pyramid vision transformer},
  author={Wang, Wenhai and Xie, Enze and Li, Xiang and Fan, Deng-Ping and Song, Kaitao and Liang, Ding and Lu, Tong and Luo, Ping and Shao, Ling},
  journal={Comput. Vis. Media},
  volume={8},
  number={3},
  pages={415--424},
  year={2022},
  publisher={Springer}
}

@inproceedings{Liu2024vmamba,
  title={{VMamba}: Visual State Space Model},
  author={Liu, Yue and Tian, Yunjie and Zhao, Yuzhong and Yu, Hongtian and Xie, Lingxi and Wang, Yaowei and Ye, Qixiang and Liu, Yunfan},
  booktitle={NeurIPS},
  year={2024}
}

@article{he2023reti,
  title={{Reti-Diff}: Illumination degradation image restoration with retinex-based latent diffusion model},
  author={He, Chunming and Fang, Chengyu and Zhang, Yulun and Ye, Tian and Li, Kai and Tang, Longxiang and Guo, Zhenhua and Li, Xiu and Farsiu, Sina},
  journal={ICLR},
  year={2025}
}

@inproceedings{bui2024meganet,
  title={{MEGANet}: Multi-scale edge-guided attention network for weak boundary polyp segmentation},
  author={Bui, Nhat-Tan and Hoang, Dinh-Hieu and Nguyen, Quang-Thuc and Tran, Minh-Triet and Le, Ngan},
  booktitle={WACV},
  pages={7985--7994},
  year={2024}
}

@inproceedings{xia2023diffir,
  title={{DiffIR}: Efficient diffusion model for image restoration},
  author={Xia, Bin and Zhang, Yulun and Wang, Shiyin and Wang, Yitong and Wu, Xinglong and Tian, Yapeng and Yang, Wenming and Van Gool, Luc},
  booktitle={ICCV},
  year={2023}
}

@article{wang2023polarization,
  title={Polarization-based camouflaged object detection},
  author={Wang, Xin and Zhang, Zhao and Gao, Jun},
  journal={Pattern Recognition Letters},
  volume={174},
  pages={106--111},
  year={2023}
}

@article{he2025unfoldir,
  title={{UnfoldIR}: Rethinking Deep Unfolding Network in Illumination Degradation Image Restoration},
  author={He, Chunming and Zhang, Rihan and Xiao, Fengyang and Fang, Chengyu and Tang, Longxiang and Zhang, Yulun and Farsiu, Sina},
  journal={arXiv preprint arXiv:2505.06683},
  year={2025}
}

@article{li2025mcod,
  title={{MCOD}: The First Challenging Benchmark for Multispectral Camouflaged Object Detection},
  author={Li, Yang and Xu, Tingfa and Bai, Shuyan and Liu, Peifu and Li, Jianan},
  journal={arXiv preprint arXiv:2509.15753},
  year={2025}
}

@article{he2025segment,
  title={Segment concealed object with incomplete supervision},
  author={He, Chunming and Li, Kai and Zhang, Yachao and Yang, Ziyun and Tang, Longxiang and Zhang, Yulun and Kong, Linghe and Farsiu, Sina},
  journal={IEEE Trans. Pattern Anal. Mach. Intell.},
  year={2025}
}

@inproceedings{pang2022zoom,
  title={Zoom in and Out: A Mixed-Scale Triplet Network for Camouflaged Object Detection},
  author={Pang, Youwei and Zhao, Xiaoqi and Xiang, Tian-Zhu and Zhang, Lihe and Lu, Huchuan},
  booktitle={CVPR},
  pages={2160--2170},
  year={2022}
}

@article{sun2025conditional,
  title={Conditional diffusion models for camouflaged and salient object detection},
  author={Sun, Ke and Chen, Zhongxi and Lin, Xianming and Sun, Xiaoshuai and Liu, Hong and Ji, Rongrong},
  journal={IEEE Transactions on Pattern Analysis and Machine Intelligence},
  year={2025},
  publisher={IEEE}
}

@article{he2025reversible,
  title={Reversible Unfolding Network for Concealed Visual Perception with Generative Refinement},
  author={He, Chunming and Xiao, Fengyang and Zhang, Rihan and Fang, Chengyu and Fan, Deng-Ping and Farsiu, Sina},
  journal={arXiv preprint arXiv:2508.15027},
  year={2025}
}

@article{he2025run,
  title={{RUN}: Reversible Unfolding Network for Concealed Object Segmentation},
  author={He, Chunming and Zhang, Rihan and Xiao, Fengyang and Fang, Chenyu and Tang, Longxiang and Zhang, Yulun and Kong, Linghe and Fan, Deng-Ping and Li, Kai and Farsiu, Sina},
  journal={ICML},
  year={2025}
}

@inproceedings{he2026nun,
  title={Nested Unfolding Network for Real-World Concealed Object Segmentation},
  author={He, Chunming and Zhang, Rihan and Tang, Longxiang and Zhang, Dingming and Zhang, Bojian and Xiao, Fengyang and Feng, Jingjia and Farsiu, Sina},
  booktitle={European Conference on Computer Vision},
  year={2026}
}

@article{he2026ride,
  title={{RIDE}: Retinex-informed decoupling for exposing concealed objects},
  author={He, Chunming and Zhang, Rihan and Zhang, Dingming and Fang, Chengyu and Tang, Longxiang and Feng, Jingjia and Xiao, Fengyang and Farsiu, Sina},
  journal={arXiv preprint arXiv:2605.15450},
  year={2026}
}

@article{land1971lightness,
  title={Lightness and retinex theory},
  author={Land, Edwin H and McCann, John J},
  journal = {Journal of the Optical Society of America},
  volume={61},
  number={1},
  pages={1--11},
  year={1971},
  publisher={Optical Society of America}
}

@inproceedings{zamir2022restormer,
  title={{Restormer}: Efficient transformer for high-resolution image restoration},
  author={Zamir, Syed Waqas and Arora, Aditya and Khan, Salman and Hayat, Munawar and Khan, Fahad Shahbaz and Yang, Ming-Hsuan},
  booktitle={Proceedings of the IEEE/CVF conference on computer vision and pattern recognition},
  pages={5728--5739},
  year={2022}
}

@article{oquab2023dinov2,
  title={{DINOv2}: Learning robust visual features without supervision},
  author={Oquab, Maxime and Darcet, Timoth{\'e}e and Moutakanni, Th{\'e}o and Vo, Huy and Szafraniec, Marc and Khalidov, Vasil and Fernandez, Pierre and Haziza, Daniel and Massa, Francisco and El-Nouby, Alaaeldin and others},
  journal={arXiv preprint arXiv:2304.07193},
  year={2023}
}

@article{jan2025c3net,
  title={{C3Net}: Context-Contrast Network for Camouflaged Object Detection},
  author={Jan, Baber and El-Maleh, Aiman H and Siddiqui, Abdul Jabbar and Bais, Abdul and Anwar, Saeed},
  journal={arXiv preprint arXiv:2511.12627},
  year={2025}
}
}

\end{document}